\documentclass[letterpaper, 10 pt, journal, twoside]{ieeetran}

\usepackage{caption}
\usepackage{amsmath,amsfonts}
\usepackage{amssymb}
\usepackage{amsthm}
\usepackage{algorithmic}
\usepackage{algorithm}
\usepackage{array}
\usepackage{booktabs}
\usepackage{textcomp}
\usepackage{multirow}
\usepackage{multicol}
\usepackage{stfloats}
\usepackage{subcaption}
\usepackage{url}
\usepackage{verbatim}
\usepackage{graphicx}
\usepackage{cite}
\usepackage{xcolor}
\usepackage{soul}
\usepackage[printonlyused]{acronym}
\usepackage{comment}
\usepackage{siunitx}
\usepackage{bm}
\usepackage{nicefrac}
\usepackage{pgfplots}
\usepackage{hyperref}
\hypersetup{
    colorlinks=false,
    linkcolor=red,
    filecolor=magenta,      
    urlcolor=cyan,
    pdftitle={CurviTrack},
    pdfpagemode=FullScreen,
}

\pgfplotsset{compat=1.14}
\usetikzlibrary{backgrounds,arrows,automata,shapes,positioning,calc,through,spy}
\pgfdeclarelayer{background}
\pgfdeclarelayer{foreground}
\pgfsetlayers{background,main,foreground}

\theoremstyle{definition}

\theoremstyle{remark}

\begin{document}
\newcommand{\usvaxis}{b}
\newcommand{\discusvaxis}{b}
\newcommand{\usvstatevect}{\mathbf{v}}
\newcommand{\usvstatevectder}{\dot{\usvstatevect}}
\newcommand{\discusvstatevect}{\mathbf{v}}
\newcommand{\discusvstatevectder}{\dot{\mathbf{v}}}
\newcommand{\phaseji}{\phi_{j,i}}
\newcommand{\Phaseji}{\Phi_{j,i}(t) }
\newcommand{\phase}{\phi}
\newcommand{\Phase}{\Phi}
\newcommand{\obsPhase}{\Phase_{j,i}(t_{obs})}
\newcommand{\ampji}{A_{j,i}}
\newcommand{\amp}{A}
\newcommand{\obsAmp}{\amp_{j,i}(t_{obs})}
\newcommand{\freqji}{f_{j,i}}
\newcommand{\freq}{f}
\newcommand{\fftdelta}{\Delta T_{FFT}}
\newcommand{\discsamplingtime}{\Delta T}
\newcommand{\identime}{t_{FFT}}

\newcommand{\usvstateworld}{\mathbf{b}}
\newcommand{\usvstateworldoscillatory}{\mathbf{b_w^O}}
\newcommand{\usvinput}{\mathbf{u_b}}
\newcommand{\dragacc}{A_d}
\newcommand{\dragaccx}{a_{x}}
\newcommand{\dragaccy}{a_{y}}
\newcommand{\dragforce}{\mathbf{F_d}}
\newcommand{\usvmass}{m}
\newcommand{\usvstatetransitionmat}{\mathbf{A}}
\newcommand{\usvstateinputmat}{\mathbf{B}}
\newcommand{\usvcovariance}{\mathbf{Q_b}}
\newcommand{\worldtobody}{~^b\mathbf{R}_w}
\newcommand{\bodytoworld}{~^w\mathbf{R}_b}
\newcommand{\coeffdragx}{K_x}
\newcommand{\coeffdragy}{K_y}
\newcommand{\droneinputvector}{\mathbf{u_d}}
\newcommand{\dronestatematrix}{\mathbf{D}}
\newcommand{\droneinputmatrix}{\mathbf{E}}
\newcommand{\dronestatevector}{\mathbf{x}}
\newcommand{\dronestatevectdes}{\overset{*}{\dronestatevector}}
\newcommand{\substatematrix}{\mathbf{D'}}
\newcommand{\subinputmatrix}{\mathbf{E'}}
\newcommand{\headingusv}{\eta_b}
\newcommand{\headinguav}{\psi_d}
\newcommand{\deltapred}{\Delta t_{p}}
\newcommand{\deltat}{dt}
\newcommand{\objfunction}{J(\dronestatevector, \droneinputvector)}
\newcommand{\errormat}{\mathbf{\Tilde{\dronestatevector}}}
\newcommand{\errpenmat}{\mathbf{S}}
\newcommand{\inputeffortvect}{\mathbf{h}}
\newcommand{\inputeffortpen}{\mathbf{T}}
\newcommand{\predhorizon}{M_p}
\newcommand{\controlhorizon}{M_c}
\newcommand{\errinz}{\Tilde{z}}
\acrodef{uav}[UAV]{Unmanned Aerial Vehicle}
\acrodef{mpc}[MPC]{Model Predictive Control}
\acrodef{nmpc}[NMPC]{Nonlinear Model Predictive Control}
\acrodef{usv}[USV]{Unmanned Surface Vehicle}
\acrodef{ekf}[EKF]{Extended Kalman Filter}
\acrodef{ukf}[UKF]{Unscented Kalman Filter}
\acrodef{lkf}[LKF]{Linear Kalman Filter}
\acrodef{fft}[FFT]{Fast Fourier Transform}
\acrodef{ode}[ODE]{ordinary differential equation}
\acrodef{dof}[DOF]{Degrees of Freedom}
\acrodef{imu}[IMU]{Inertial Measurement Unit}
\acrodef{gnss}[GNSS]{Global Navigation Satellite System}
\acrodef{uvdar}[UVDAR]{UltraViolet Direction And Ranging}
\acrodef{rmse}[RMSE]{Root Mean Square Error}
\acrodef{lidar}[LiDAR]{Light Detection and Ranging}
\acrodef{mekf}[MEKF]{Multiplicative Extended Kalman Filter}
\acrodef{indi}[INDI]{Incremental Non-linear Dynamic Inversion}
\acrodef{ocp}[OCP]{Optimal Control Problem}

\newcommand{\raw}[1]{\textcolor{orange}{#1}}
\newcommand{\reffig}[1]{figure~\ref{#1}}

\newcommand{\RF}[1]{#1}
\newcommand{\RS}[1]{#1}

\title{Sensorless State Estimation and Control for Agile Cable-Suspended Payload Transport by Quadrotors}

\author{Ana Maria P. S. Nascimento, \IEEEmembership{Member,~IEEE}, Augusto V. Sales, \IEEEmembership{Student Member,~IEEE}, Antonio Marcus N. Lima, \IEEEmembership{Senior Member,~IEEE} and Tiago P. Nascimento, \IEEEmembership{Senior Member,~IEEE}% <-this % stops a space
%\author{Anonimous Authors
    \thanks{Manuscript received: XXXXXX XXth, 2025; Revised XXXXXXX XXth, 2026; Accepted XXXXXXX XXth, 2026.}
    % \thanks{This paper was recommended for publication by Editor Giuseppe Loianno upon evaluation of the Associate Editor and Reviewers' comments.}
	% \thanks{This work has been supported by the National Council for Scientific and Technological Development – CNPq, by the National Fund for Scientific and Technological Development – FNDCT, by the Ministry of Science, Technology and Innovations – MCTI from Brazil under research project No. 401097/2025-0, 443960/2024-0, and 304551/2023-6.}% <-this % stops a space
	% \thanks{A. M. P. S. Nascimento and A. M. N. Lima are with the Department of Electrical Engineering, Universidade Federal de Campina Grande, Brazil.}
    % \thanks{T. Nascimento and A. V. Sales are with the Department of Computer Systems, Universidade Federal da Paraíba, Brazil.}
    \thanks{Digital Object Identifier (DOI): see top of this page.}
}
%\markboth{IEEE Robotics and Automation Letters. Preprint Version. Accepted XXXXX, 2026}
\markboth{ArXiv. Preprint Version. Accepted XXXXX, 2026}
{Nascimento \MakeLowercase{\textit{et al.}}: Sensorless Estimation and Control for Agile Cable-Suspended Payload Transportation}  
\maketitle

\begin{abstract}
This work proposes a novel control and estimation approach for aerial manipulation of a cable-suspended load using Unmanned Aerial Vehicles (UAVs). Common approaches in the state of the art have practical limitations, relying on direct load measurements and Lagrangian methods for dynamic modeling. The lack of a straightforward dynamic model of the system led us to propose adopting the Udwadia–Kalaba method to explicitly incorporate the cable's geometric constraints. This formulation allowed for the consistent derivation of the tension force and its direct integration into the \ac{nmpc} prediction model. Additionally, we propose a sensorless load state estimation based on the same geometric constraints. Results from real-robot experiments demonstrated that the explicit inclusion of load dynamics in the optimization problem significantly reduces trajectory-tracking errors and yields better overall performance compared to strategies based on incomplete models.
\end{abstract}


\section{Introduction}

\begin{figure}[!t]
    \centering
    \includegraphics[width=0.4\textwidth]{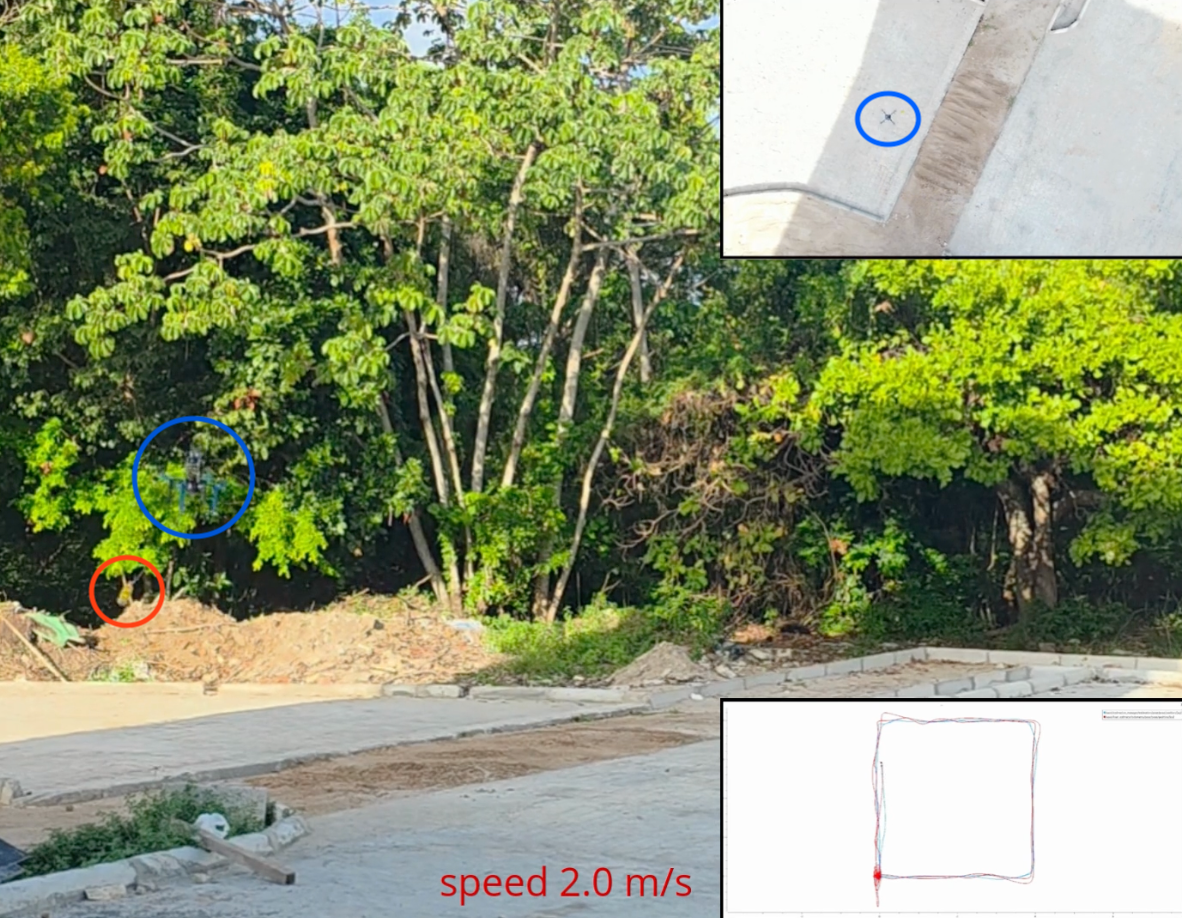}
    \caption{\ac{uav} at 2 m/s transporting a 0.5 kg payload through a 1 m length cable.}
    \label{fig:uav_exemplo}
\end{figure}

\IEEEPARstart{T}{he} multirotor \acp{uav} have played an increasingly important role in industrial, security, inspection, and rescue applications \cite{villa2020survey}. One particularly promising, yet technically challenging, application is the transportation of cargo by \acp{uav} (Figure \ref{fig:uav_exemplo}). This problem has been extensively studied in recent literature \cite{LEE2022844, estevez2024drones}, both in the context of cargo rigidly coupled to the vehicle and in the transport of cargo suspended by cables. This work focuses primarily on the latter case, where a quadrotor transports cargo connected by an inelastic cable.

The main challenge in transporting suspended cargo using \acp{uav} lies in the pendulum dynamics induced by the cable, which can degrade control performance and compromise flight stability \cite{Sreenath2013Geometric}. These oscillations not only hinder accurate trajectory tracking but also introduce additional forces into the system, resulting in strongly coupled and nonlinear dynamics. Developing robust and predictive control strategies is therefore essential to mitigate these effects and ensure safe and efficient operation \cite{Guo2018ControllingAQ}.

In control theory, \ac{nmpc} optimizes future control actions for a nonlinear system over a prediction horizon by minimizing a cost function subject to state and input constraints. Its ability to handle nonlinearities and constraints explicitly makes it well suited to controlling \acp{uav} such as quadrotors \cite{Tiago2019}. In this work, we make the following contributions for quadrotor-based payload transportation:
\begin{itemize}
    \item An \ac{nmpc} approach for quadrotor payload transportation based on the Udwadia--Kalaba formulation, integrating cable dynamics and external-force estimation into the prediction model.
    \item A payload state-estimation method for a single quadrotor that does not require additional sensors on the cable or payload.
    \item Real-world experimental validation with a quadrotor carrying a slung payload, demonstrating performance in outdoor conditions with varying illumination and random wind disturbances.
\end{itemize}

\section{Related Work}

Existing approaches in the literature use simplified models of the cable-load system and assume ideal constraints. Furthermore, they rely on high-frequency load and cable measurements for closed-loop control, typically requiring additional sensors mounted on the load, such as reflective markers for a motion capture system, or on the quadrotor, such as downward-facing cameras, cable tension sensors, and cable direction sensors \cite{Sun2025}.

Accurate system modeling is essential for developing effective control strategies. Early studies investigated approaches such as geometric control \cite{Sreenath2013Geometric} and adaptive control \cite{Dai2014Adaptive} for a quadrotor \ac{uav} transporting a cable-suspended payload, assuming the cable behaved as a rigid link. Later, Kotaru et al. \cite{Kotaru2017DynamicsAC} studied the dynamics and control of a quadrotor carrying a load suspended by an elastic cable modeled with spring stiffness and damping. In 2018, Guo et al. \cite{Guo2018ControllingAQ} designed an optimal control system for a quadcopter transporting a cable-suspended load through a narrow window, using Low Equivalent System (LOES) identification for the system model.

Recent studies have examined the modeling and control of quadrotor--payload systems. Sun et al. \cite{jihaosun2021} modeled the payload as a point mass connected to a \ac{uav} by a rigid cable, which simplifies the equations of motion. Lv et al. \cite{lv2022finite} extended this framework by modeling cable flexibility to capture oscillations and vibrations that can degrade performance. Chang et al. \cite{chang2023adaptive} proposed a model that includes damping and time-varying parameters to adapt to changing operating conditions. Belguith et al. \cite{belguith2023modeling} investigated payload aerodynamics and its interaction with the \ac{uav} structure, yielding a more complete model. Goodman et al. \cite{goodman2023geometric} showed that Lie-group-based geometric control can improve trajectory tracking accuracy.

Load oscillations during transport remain a major challenge and motivate robust control strategies. Urbina-Brito et al. \cite{UrbinaBrito2021Predictive} proposed a Model Predictive Control (MPC) scheme for quadrotors transporting suspended loads that accounts for load dynamics, full 3D vehicle motion, and attitude dynamics. In the same year, Li et al. \cite{Li2021PCMPC} introduced Perception-Constrained Model Predictive Control (PCMPC), which incorporates perception constraints alongside nonlinear dynamics and actuator limits. Later, Belguith et al. \cite{belguith2023modeling} applied a sliding-mode controller for oscillation compensation and stability. Outeiro et al. \cite{Outeiro2023Control} proposed a control architecture for transporting suspended loads with uncertain mass, addressing both load dynamics and parametric uncertainty. Li et al. \cite{Li2023AutoTrans} proposed AutoTrans, an integrated planning and control framework for aerial payload transport that combines real-time spatiotemporal planning with an adaptive \ac{nmpc} guidance layer and hierarchical disturbance compensation, including an online external force estimator, although it relies on IMU data mounted on the cable.

In 2024, Panetsos et al. \cite{panetsos2024nmpc} implemented an \ac{nmpc} scheme for disturbance anticipation and load-transport optimization, using vision-based state estimates. Wang et al. \cite{Wang2024ImpactAware} proposed an impact-aware planning and control framework for aerial robots with suspended payloads, targeting impacts induced by cable mode transitions (taut/slack). More recently, Sun et al. \cite{Sun2025} presented a trajectory-centric framework that solves the whole-body kinodynamic motion planning problem online while accounting for dynamic coupling and physical constraints between the \acp{uav} and the suspended load.

\section{Preliminaries}

We first define the main variables used in this work, expressed in the \textit{world frame} and in SI units. In our model, $\mathbf{p}_Q \in \mathbb{R}^3$, $\dot{\mathbf{p}}_Q \in \mathbb{R}^3$, and $\ddot{\mathbf{p}}_Q \in \mathbb{R}^3$ denote the \ac{uav} position, velocity, and acceleration, respectively, and $\mathbf{q}_Q \in \mathbb{S}^3$ denotes its attitude quaternion. Part of the dynamic formulation is expressed in the \textit{robot frame}, where forces and torques are applied. Accordingly, $\mathbf{v}_Q \in \mathbb{R}^3$, $\bm{\omega} \in \mathbb{R}^3$, and $\dot{\bm{\omega}} \in \mathbb{R}^3$ denote the linear velocity, angular velocity, and angular acceleration of the \ac{uav} in the robot frame, respectively; $\bm{\dot{\delta}} \in \mathbb{R}^4$ denotes the propeller rotational speeds; $T = \sum_{i=1}^4 T_i$ denotes the total thrust; and $\bm{\tau} \in \mathbb{R}^3$ denotes the torque vector for roll, pitch, and yaw. Finally, the models in Sections~\ref{sec:model} and \ref{sec:force_est} are jointly used as the prediction model for the proposed \ac{nmpc} in Section~\ref{sec:nmpc}.
\begin{figure}[!h]
    \centering
    \includegraphics[width=0.35\textwidth]{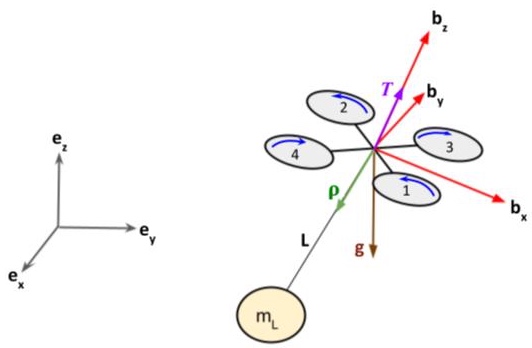}
    \caption{Schematic of a \ac{uav} carrying a payload via a cable with axis convention.}
    \label{fig:uav}
\end{figure}

\subsection{Cable-Suspended Payload Aerial System Dynamic Model}
\label{sec:model}

Let us now also consider a dynamic system consisting of a \ac{uav} carrying a payload of mass $m_L$ by means of a rigid, inextensible, and massless cable of length $L_c$, assumed to be always taut throughout the operation, as described in Fig. \ref{fig:uav}. At the same time, let $\mathbf{p}_L\in \mathbb{R}^3$, and $\mathbf{\dot{p}}_L\in \mathbb{R}^3$ be the position of the payload, and the linear velocity of the payload, respectively, all expressed in the \textit{world frame}. Furthermore, the direction of the cable is represented by the unit vector $\bm{\rho} \in \mathbb{S}^2$, oriented from the quadcopter to the payload. In this UAV-payload system, $\bm{R}(\mathbf{q}_Q)$ is the quaternion rotation that transforms vectors from the robot's frame to the world frame, the operation $\otimes$ is the Hamilton multiplication (quaternion multiplication), $m_Q$ is the total mass of the \ac{uav}, $\hat{\mathbf{e}}_z$ is the vertical unit vector in the world frame, $\hat{\mathbf{b}}_z$ is the vertical unit vector (in the robot frame), $\hat{\mathbf{e}}_z$ is the vertical unit vector in the world frame, $\mathbf{I}_Q$ is the inertia matrix of the \ac{uav} body (robot frame), and the gravity vector is such that $\bm{g}=[0, 0, -9.8]^\top$. Finally, let us also initially consider that the quadcopter-payload system is not bounded by a holonomic constraint imposed by the cable. Then, the overall model of the UAV-payload system for cable transport is given by
\begin{align}
    & \mathbf{\dot{p}}_Q = \bm{R}(\mathbf{q}_Q)  \mathbf{v}_Q \label{eq:vel},  \\
    & m_Q  \mathbf{\ddot{p}}_Q = \bm{R}(\mathbf{q}_Q)(T \hat{\mathbf{b}}_z) - m_Q  \bm{g}  \hat{\mathbf{e}}_z + \bm{F}_L \label{eq:force},\\
    & \dot{\mathbf{q}}_Q = \frac{1}{2}\mathbf{q}_Q \otimes \begin{bmatrix} 0 \\ \bm{\omega} \end{bmatrix} \label{eq:quater},\\
    &  \mathbf{I}_Q\dot{\bm{\omega}} = -\bm{\omega} \times \mathbf{I}_Q\bm{\omega} + \mathbf{\tau} 
    \label{eq:torque},\\
    &  \mathbf{\dot{p}}_L = \frac{d \mathbf{p}_L}{d t} \label{eq:velLoad}, \\ 
    &  m_L  \mathbf{\ddot{p}}_L = - m_L \bm{g} \hat{\mathbf{e}}_z  - \bm{F}_L \label{eq:forceLoad}.
\end{align}
% \noindent where 

Note that, in the above dynamic model, the components for the orientation and angular velocity of the load are missing. This is because, without sensors, it is only possible to calculate these components if we have a multi-robot system that acts directly influencing it \cite{Sun2025}. Thus, the full state vector can be written as follows
\begin{equation}
    \bm{\xi}_k = [\bm{\xi}_{Q,k}, \; \bm{\xi}_{L,k}]^\top, 
\end{equation}
where the state of the quadrotor is such that $\bm{\xi}_{Q,k} = [\mathbf{p}_{Q,k}^\top, \; \mathbf{\dot{p}}_{Q,k}^\top, \; \mathbf{q}_{Q,k}^\top, \; \bm{\omega}_{Q,k}^\top]^\top$.

Additionally, the force and torque vector can also be expressed compactly as
\begin{align}
\begin{bmatrix}
T \\ \bm{\tau}
\end{bmatrix} = 
\mathbf{G_1} \bm{u} + \mathbf{G_2} \ddot{\bm{\delta}} +  \mathbf{G_3} (\bm{\omega}) \dot{\bm{\delta}},
\end{align}
where $\bm{u} = c_t \ddot{\bm{\delta}}^2$ is the individual thrust vector of the engines, with $c_t$ being the thrust coefficient. The matrices $\mathbf{G_1}$, $\mathbf{G_2}$, and $\mathbf{G_3}$ describe, respectively, the input coupling, the inertia of the rotors, and the resultant of the gyroscopic effect between the rotors and the \ac{uav} body, such that

\begin{small}
\begin{align}
\mathbf{G_1} = \begin{bmatrix}
1 & 1 & 1 & 1 \\
-l_d \sin \beta  & l_d \sin \beta  & l_d \sin \beta  & -l_d \sin \beta  \\
l_d \sin \beta  & -l_d \sin \beta  & -l_d \sin \beta  & l_d \sin \beta  \\
-c_q / c_t & -c_q / c_t & c_q / c_t & c_q / c_t
\end{bmatrix}, 
\label{eq:g1}
\end{align}
\end{small}
\begin{scriptsize}
\begin{align}
\mathbf{G_2} = \begin{bmatrix}
0 & 0 & 0 & 0 \\
0 & 0 & 0 & 0 \\
0 & 0 & 0 & 0 \\
-I_p & -I_p & I_p & I_p
\end{bmatrix},
\end{align}
\end{scriptsize}
\noindent and
\begin{scriptsize}
\begin{align}
\mathbf{G_3} = \begin{bmatrix}
0 & 0 & 0 & 0 \\
-I_p \omega_\theta & -I_p \omega_\theta & I_p \omega_\theta & I_p \omega_\theta \\
I_p \omega_\phi & I_p \omega_\phi & -I_p \omega_\phi & -I_p \omega_\phi \\
0 & 0 & 0 & 0
\end{bmatrix},
\end{align}
\end{scriptsize}\noindent where $l_d$ is the distance from the center of mass to the rotor [m], $\beta$ is the angle between $\hat{\bm{b}}_x$ and the arm of rotor 3, $c_q$ the torque coefficient, and $I_p$ the rotor's inertia around $\hat{\bm{e}}_z$. Furthermore, the matrices above can be derived from the convention shown in Fig. \ref{fig:uav}. It is worth mentioning that both matrices $\mathbf{G_2}$ and $\mathbf{G_3}$ are disregarded for the \ac{nmpc} controller design.

\subsection{External Force Estimator}
\label{sec:force_est}

In order to estimate the external force exerted by the payload-cable system on the \ac{uav}, let us now consider that the payload and the vehicle are treated as rigid bodies coupled by a holonomic constraint imposed by the cable. Then, the generalized coordinates are defined as
%\begin{equation}
$\bm{P} = \begin{bmatrix}
\mathbf{p}_Q, \;
\mathbf{p}_L
\end{bmatrix}^\top \in \mathbb{R}^6$.
%\end{equation}

The mass matrix of the free system is given by
\begin{equation}
\bm{M} =
\begin{bmatrix}
m_Q \bm{I}_3 & 0 \\
0 & m_L \bm{I}_3
\end{bmatrix}.
\end{equation}

The generalized force vector of the free system can be written as
\begin{equation}
\bm{F}_s =
\begin{bmatrix}
\bm{R}(\mathbf{q}_Q) (T\hat{\bm{b}}_z) - m_Q \bm{g} \hat{\bm{e}}_z \\
- m_L \bm{g} \hat{\bm{e}}_z
\end{bmatrix},
\end{equation}
\noindent and the translational dynamics of the system are described as being $\bm{M} \ddot{\bm{P}} = \bm{F}_s$.

In this system, the cable is modeled as rigid, inextensible, and of constant length, which imposes the following holonomic motion constraint
\begin{equation}
\zeta(\bm{P}) = \|\mathbf{p}_L - \mathbf{p}_Q \|^2 - L_c^2 = 0.
\label{eq:constraint}
\end{equation}

By deriving this constraint twice in time, we obtain the acceleration constraint as being
\begin{equation}
\bm{\rho}_N^\top (\ddot{\mathbf{p}}_L - \ddot{\mathbf{p}}_Q)
=
- \frac{\| \dot{\mathbf{p}}_L - \dot{\mathbf{p}}_Q \|^2}{L_c},
\end{equation}
\noindent with
\begin{align}
    & \bm{\rho}_N = \frac{\bm{\rho}}{\|\bm{\rho} \|} = \frac{\mathbf{p}_L - \mathbf{p}_Q}{\| \mathbf{p}_L - \mathbf{p}_Q \|}, \label{eq:rho_norm}, \\
    & \bm{\rho} = \mathbf{p}_L - \mathbf{p}_Q \label{eq:rho}, \\
    & L_c = \|\bm{\rho} \| = \| \mathbf{p}_L - \mathbf{p}_Q \| \label{eq:cabe_comp}, \\
    & \bm{V}_R = \dot{\mathbf{p}}_L - \dot{\mathbf{p}}_Q \label{eq:velocidadeAngularBodyFrame},
\end{align}
\noindent where $\bm{V}_R$ is the relative velocity of the payload with respect to the \ac{uav}.

Once the UAV–load system model is obtained without explicitly imposing the cable constraint, the challenge lies in extending it to account for this constraint, which is nontrivial under the standard Newton–Euler formulation due to its complexity and configuration-specific nature. While the Lagrange–d’Alembert principle is widely used in the literature to model drone–cargo systems with eight degrees of freedom, including underactuated ones, it relies on the assumption of ideal constraints and can become cumbersome when multiple configurations are considered \cite{villa2020survey,Wang2024ImpactAware}. In contrast, the Udwadia–Kalaba formulation \cite{Udwadia2002} provides a more recent and explicit approach for deriving system dynamics, is applicable to both compact and augmented systems, and offers a simpler extension of Newton–Euler models. Consequently, this is particularly suitable for practical \ac{uav} applications that employ nonlinear and predictive control strategies.

Thus, we can then apply the constraint seen in the equation \eqref{eq:constraint} using the method proposed by Udwadia–Kalaba, as
\begin{equation}
\bm{A}(\bm{P})\ddot{\bm{P}} = b(\bm{P},\dot{\bm{P}}),
\end{equation}
with $\bm{A} = \begin{bmatrix}
\bm{\rho}^\top, \; -\bm{\rho}^\top
\end{bmatrix}
\in \mathbb{R}^{1 \times 6}$, and
$b = - \frac{\| \dot{\mathbf{p}}_L - \dot{\mathbf{p}}_Q \|^2}{L_c}$.

Accordingly, the generalized constraint force ($\bm{F}_{RS}$) is given by
\begin{equation}
\bm{F}_{RS} = \bm{M}^{1/2} \bm{A}^+ \left(b - \bm{A}\bm{M}^{-1}\bm{F}_s \right),
\end{equation}
where $\bm{A}^+$ denotes the Moore–Penrose pseudo-inverse of the matrix $\bm{A}\bm{M}^{-1/2}$.

Note here that the restraining force acts exclusively along the direction of the cable. Therefore, $\bm{F}_{RS}$ can be physically interpreted as the tension force in the cable ($\bm{F}_L$), such that
\begin{equation}
\bm{F}_{RS} = \bm{F}_L =
\begin{bmatrix}
\bm{\rho}_N \lambda, \;
-\bm{\rho}_N \lambda
\end{bmatrix}^\top,
\end{equation}
\noindent where $\lambda \geq 0$ is the magnitude of the tension in the cable, which must be estimated.

Note also that the total dynamics of the system (\ac{uav} + cable + load) with constraint is now given by
\begin{equation}
\bm{M} \ddot{\bm{P}} = \bm{F}_S + \bm{F}_L.
\end{equation}

Which, in turn, we can rewrite this equation such that
\begin{align}
m_Q \ddot{\mathbf{p}}_Q &= \bm{R}(\mathbf{q}_Q)(T \hat{\mathbf{b}}_z) - m_Q \bm{g} \hat{\mathbf{e}}_z + \bm{F}_L, \label{eq:uav-cabo}\\
m_L \ddot{\mathbf{p}}_L &= - m_L \bm{g} \hat{\mathbf{e}}_z - \bm{F}_L \label{eq:cabo-carga}.
\end{align}

In the Udwadia-Kalaba formulation, the cable tension force ($\bm{F}_L$) is not explicitly modeled, but is instead derived from the dynamics of the free system. Specifically, this force is identified as the minimal correction required to modify the free system dynamics, which in turn occur in the absence of a cable, such that the trajectory strictly satisfies the specific holonomic constraint.

Let us now initially consider the free accelerations of the quadcopter and the payload, obtained from equations (\ref{eq:uav-cabo}) and (\ref{eq:cabo-carga}). Then, we have that
\begin{align}
\ddot{\mathbf{p}}_Q^{\,\text{free}} &= \frac{1}{m_Q} ( \bm{R}(\mathbf{q}_Q)T \hat{\mathbf{e}}_z) - \bm{g}  \hat{\mathbf{e}}_z, \label{eq:pq_free}\\
\ddot{\mathbf{p}}_L^{\,\text{free}} &= - \bm{g} \hat{\mathbf{e}}_z
\label{eq:pl_free}.
\end{align}

The relative acceleration along the direction of the cable ($\bm{a}_R$) is given by
\begin{equation}
\bm{a}_R = \bm{\rho}_N^\top \left( \ddot{\mathbf{p}}_L^{\,\text{free}} - \ddot{\mathbf{p}}_Q^{\,\text{free}} \right).
\label{eq:free_rel_acc}
\end{equation}

However, the holonomic constraint of the cable imposes that the relative acceleration along $\bm{\rho}$ be satisfied, such that
\begin{equation}
\bm{\rho}_N^\top (\ddot{\mathbf{p}}_L^{\,\text{free}} - \ddot{\mathbf{p}}_Q^{\,\text{free}}) = - \frac{1}{L_c}\|\dot{\mathbf{p}}_L - \dot{\mathbf{p}}_Q\|^2.
\label{eq:constraint_acc}
\end{equation}

It is observed that the free dynamics generally do not satisfy this constraint. Consequently by referring to \eqref{eq:velocidadeAngularBodyFrame}, we define a scalar discrepancy function ($\bm{E}$) as follows
\begin{equation}
\bm{E} = \bm{a}_R + \frac{1}{L_c}\|\bm{V}_R\|^2,
\end{equation}
which quantifies how much the free dynamics violates the cable's holonomic constraint.

The Udwadia-Kalaba formulation demonstrates that the violation of the holonomic constraint is effectively neutralized by a cable tension force ($\bm{F}_L$) acting along the constraint's direction. This force induces corrections to the quadrotor and payload accelerations required to ensure that
\begin{equation}
\ddot{\mathbf{p}}_Q = \ddot{\mathbf{p}}_Q^{\,\text{free}} + \frac{1}{m_Q} \lambda \bm{\rho}_N, \quad
\ddot{\mathbf{p}}_L = \ddot{\mathbf{p}}_L^{\,\text{free}} - \frac{1}{m_L} \lambda \bm{\rho}_N. 
\label{eq:acc_pL}
\end{equation}

Thus, by substituting these expressions into the acceleration constraint \eqref{eq:constraint_acc}, the magnitude of the cable tension ($\lambda$) is found to be
\begin{equation}
\lambda = \frac{
\bm{\rho}_N^\top \left( \ddot{\mathbf{p}}_L^{\,\text{free}} - \ddot{\mathbf{p}}_Q^{\,\text{free}} \right) + \frac{1}{L_c}\|\dot{\mathbf{p}}_L - \dot{\mathbf{p}}_Q\|^2}{\frac{1}{m_Q} + \frac{1}{m_L}}.
\label{eq:lambda_uk}
\end{equation}

The resulting tension force acting on the system ($\bm{F}_L$) is given by
\begin{equation}
\bm{F}_L = \lambda \bm{\rho}_N.
\end{equation}

Notably, the Udwadia-Kalaba method allows for the explicit and consistent calculation of the cable tension force, ensuring strict compliance with the holonomic constraint while circumventing the numerical singularities often encountered in steady-state flight regimes. Within this framework, tension is not treated as an independent state or an arbitrary external input; rather, it emerges analytically as the force required to maintain compatibility between the system dynamics and the geometric inextensibility of the cable.

\section{Proposed System}

\begin{figure*}[!t]
    \centering
    \includegraphics[width=1.0\linewidth]{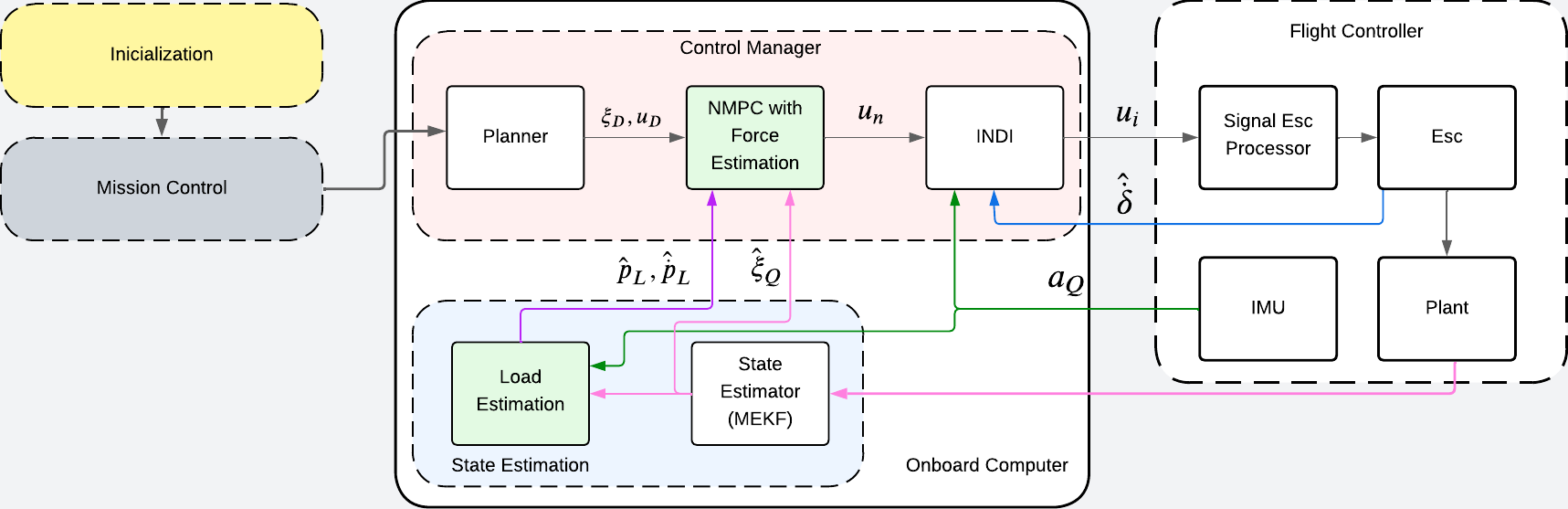}    
    \caption{System's Architecture Diagram.}
    \label{fig:arquitetura}
\end{figure*}

The \ac{uav} navigation system used here was developed by our team\footnote{suppressed for blind revision},%\url{https://github.com/LASER-Robotics/laser_uav_system}}, 
and it is implemented in a quadrotor's onboard computer. As mentioned above, our work proposes a new NMPC-based control formulation and a payload state estimator, which are represented by the \textbf{green blocks} in Figure \ref{fig:arquitetura}.

Within this diagram, we can note as green blocks the proposed control system, which is composed of an \ac{nmpc} with an external force estimation using the Udwadia–Kalaba formulation \cite{Udwadia2002} and the load state estimation. Besides the \ac{nmpc}, our \ac{uav} also uses an \ac{indi} controller from our original system.

\section{Motion Control System}
\label{sec:nmpc}

In our work, we propose an \ac{nmpc} to improve the UAV's ability to track a desired trajectory despite payload-induced disturbances while traveling at high speeds. As above mentioned, the UAV-payload system was modeled according to the Udwadia–Kalaba formulation and used as the motion controller's prediction model. The prediction model also uses the model of the external force ($\bm{F}_L$) induced by the load with the Udwadia–Kalaba formulation, which in turn is treated as a disturbance within the prediction horizon.

Let us then first consider the vectors $u_D$, $u_N$, and $u_I$ in Fig. \ref{fig:arquitetura}, where these represent the desired motor thrust, the \ac{nmpc} control input, and the INDI control input, respectively. Note that all three vectors have the same structure in which $u_s = [T_1, T_2, T_3, T_4]$, with $u_s \in [u_D, u_N, u_I]$ and where $T_i$ are the individual thrusts of each motor of the quadcopter. Thus, the \ac{nmpc} will compute a control input in the same format as this vector, now denoted by $u_N$.

The \ac{nmpc} is an optimal controller whose problem formulation is expressed as follows
\begin{footnotesize}
\begin{equation}
\begin{split}
\bm{u}_N =\operatorname*{argmin}_{u} \sum_{j = 0}^{N-1} (\left\lVert \hat{\bm{\xi}}_{Q,j} - \bm{\xi}_{D_Q,j} \right\rVert^2_{S_Q} + \left\lVert \hat{\bm{\xi}}_{L,j} - \bm{\xi}_{D_L,j} \right\rVert^2_{S_L} +\\
+ \left\lVert \bm{u}_j - \bm{u}_{D,j} \right\rVert^2_{C})
+\left\lVert \hat{\bm{\xi}}_N - \bm{\xi}_{D,N} \right\rVert^2_{S_N}
\label{eq:func_min}
\end{split}
\end{equation}
\end{footnotesize}
\noindent subject to
\begin{align}
{\bm{\xi}_{j+1}} &= \bm{f}(\bm{\xi_j},\bm{u_j}), & \bm{\xi}_0 = \bm{\xi}_{init} \label{eq:dyn_model_ocp} \\
\bm{\omega} &\in [\omega_{\text{min}} , \omega_{\text{max}}], & \bm{u} \in [\bm{u}_{\text{min}},\bm{u}_{\text{max}}]
\end{align}
\noindent where $j$ corresponds to the discrete instant of the prediction horizon, $\bm{f}$ corresponds to equations \eqref{eq:vel}-\eqref{eq:forceLoad}, $\bm{\xi}_{D_Q,k}$ and $\bm{\xi}_{D_L,k}$ are the reference state vector for the UAV and Load, respectively, $\hat{\bm{\xi}}_{Q,k}$ and $\hat{\bm{\xi}}_{L,k}$ are the current state vector for the UAV and Load, respectively, $\bm{u}_k$ is the control input (the thrust vector of the four thrusters), N is the system prediction horizon, $\hat{\bm{\xi}}_{N}$ is the drone state vector at the end of the prediction horizon, and where the input constraints $\bm{u}_{\text{min}}$ and $\bm{u}_{\text{max}}$ are the minimum and maximum thrust limits of the individual propellers, in which, for drone control, $\bm{u}_{\text{min}}$ is typically $\mathbf{0}$ and $\bm{u}_{\text{max}}$ is the maximum thrust of each propeller. Additionally, the matrices $\bm{S}_Q$, $\bm{S}_L$, and $\bm{C}$ are weight matrices that map the UAV's state, the payload's state, and the control effort, respectively, in which $\bm{S}$ can also be expressed as
$S=\text{diag}(S_p, S_{\dot{p}}, S_q, S_{\dot\delta})$, with $S_N = [S_Q,S_L]$.

Rotational dynamics around the yaw axis are inherently slower than roll and pitch responses. To optimize tracking, the attitude error is decomposed to apply distinct weights in the cost matrix $\mathbf{S}_Q$. The orientation error is defined by the error quaternion $\mathbf{q}_e$, which represents the rotation required to align the current state $\mathbf{q}$ with the reference $\mathbf{q}_D$:
\begin{equation}
    \mathbf{q}_e = \mathbf{q} \otimes \mathbf{q}_D^{-1}
\end{equation}

Finally, $\bm{q}_e$ is decomposed \textbf{into the rotational errors} of the $X$, $Y$, and $Z$ axes, based on the projection of the error vector as follows
\begin{equation} 
\label{erro_rotacional_decomposto}
\begin{bmatrix}
 q_x \\
 q_y \\
 q_z
\end{bmatrix} = \frac{1}{\sqrt{q_{ew}^2 + q_{ez}^2}} \begin{bmatrix}         q_{ew}q_{ex} + q_{ez}q_{ey} \\
 q_{ew}q_{ey} - q_{ez}q_{ex} \\
 q_{ez}
\end{bmatrix} 
\end{equation}

\section{Load State Estimator}
\label{sec:load_est}

In our system, the quadrotor estimates the quadrotor states by using an \ac{mekf} (see Fig. \ref{fig:arquitetura}) which uses the full model of the quadrotor (equations \ref{eq:vel}-\ref{eq:torque}). However, we also need to estimate the position of the load $\mathbf{p}_L \in \mathbb{R}^3$ and the linear velocity of the load $\mathbf{\dot{p}}_L \in \mathbb{R}^3$, both in the world frame. So let us consider the state vector that we want to estimate as
\begin{align}
    \bm{\xi}_{L,k} = \begin{bmatrix} \mathbf{p}_{L,k}^\top,\;
                        \mathbf{\dot{p}}_{L,k}^\top,\;
                        \bm{\rho}_k^\top,\;
                        \bm{\omega}_{L,k}^\top
    \end{bmatrix}^\top \in \mathbb{R}^{12},
\end{align}
where $k$ corresponds to the discrete instar of time using Forward Euler discretization,  $\mathbf{p}_{L,k} \in \mathbb{R}^3$ is the position of the load in the world frame, $\mathbf{\dot{p}}_{L,k} \in \mathbb{R}^3$ is the linear velocity of the load in the world frame, and $\bm{\omega}_{L,k} \in \mathbb{R}^3$ is the angular velocity of the cable in the world frame, in an instant $k$.

Here, $\bm{\rho}_k \in \mathbb{S}^2$ is the direction of the cable in the world frame, in an instant $k$, that points from the quadrotor to the load. We also consider that
\begin{equation}
    \boldsymbol{\rho} =\frac{\mathbf p_L - \mathbf p_Q}{||\mathbf p_L - \mathbf p_Q||}.
\end{equation}

This assumption results in a geometric constraint such that
\begin{equation}
    \mathbf p_L = \mathbf p_Q + L\boldsymbol{\rho} \quad\Longleftrightarrow\quad
\mathbf p_Q = \mathbf p_L - L\boldsymbol{\rho}.
\end{equation}

Before anything else, keep also in mind that $||\bm{\rho}||=1$, $\bm{\rho}^\top\bm{\omega}_L=0$, and that $\bm{\rho}$ is defined quad$\rightarrow$load. Thus, here we will propose an \ac{ekf} to estimate the load's states. Our EKF will use the drone's IMU-acceleration expressed in the body frame as the control input to the filter. Knowing that $a_Q$ is the specific force (gravity excluded) that comes from this sensor, we can find the true acceleration in \textbf{the world frame} as
\begin{equation}
    \mathbf{\ddot{p}}_{Q,k} = \mathbf{R}(\mathbf{q_{Q,k}})\mathbf{a}_{Q,k} + \bm{g},
\end{equation}
where $\mathbf{\ddot{p}}_{Q,k}$ is the linear acceleration of the quadrotor in the world frame in which $\bm{u}_k = \mathbf{\ddot{p}}_{Q,k}$.

Meanwhile, the quadrotor estimated states will be used here as sensor measurement inputs during the update step. Thus, the initial inputs for our proposed EKF are $\bm{\xi}_{L,0}$ with $\bm{\rho}_0=[0,0,-1]^\top$, and $\bm{P}_{L,0}=\bm{I}_{12}*0.1$ as the initial covariance matrix, with $\bm{I}_{12}$ is a 12-order identity matrix. Note that in order for our model to work, we need persistent excitation over time, or $(\bm{I}_3-\boldsymbol{\rho}\boldsymbol{\rho}^\top)(\mathbf{\ddot{p}}_{Q}-\bm{g})\neq 0$, where $\bm{I}_3$ is a 3-order identity matrix. Thus, when the \ac{uav} is in hover mode, $\mathbf{\ddot{p}}_{Q}\approx \bm{g} \Rightarrow f_{\omega\rho}\approx0$ and the system loses observability.

\subsection{Extended Kalman Filter for Load Estimation}

Our proposed EKF uses, then, as inputs the load state $\bm{\xi}_{L,k}$, covariance matrix $P_{L,k}$, control input $\bm{u}_k=\mathbf{\ddot{p}}_{Q,k}$ that comes from the UAV's IMU, and measurement $\bm{z}_k = \hat{\bm{\xi}}_{Q,k}$ that comes from the robot's MEKF. We also need to calculate the tension acceleration in the cable, which is
\begin{equation}
    t_{\alpha,k} = -\boldsymbol{\rho}_{k}^\top(\mathbf{\ddot{p}}_{Q,k} - \bm{g} )+L||\bm{\omega}_{L,k}||^2.
\end{equation}

Then, let us perform the raw propagation by computing the predicted state $\hat{\bm{\xi}}_{L,k+1|k} = f_p(\bm{\xi}_{L,k}, \bm{u}_k)$ using your discrete-time Prediction Model equations. Thus,
\begin{footnotesize}
\begin{align}
 f_p(\bm{\xi}_{L,k}, u_{k}) = \begin{bmatrix} \hat{\mathbf{p}}_{L,k+1|k} +\\
 \hat{\dot{\mathbf{p}}}_{L,k+1|k}\\
 \hat{\bm{\rho}}_{k+1|k}\\
 \hat{\bm{\omega}}_{L,k+1|k}
    \end{bmatrix} = \begin{bmatrix}
    \mathbf{p}_{L,k} + t_s \mathbf{\dot{p}}_{L,k} + \frac{t_s^2}{2}\mathbf{\ddot{p}}_{L,k},\\    
    \mathbf{\dot{p}}_{L,k} + t_s \mathbf{\ddot{p}}_{L,k},  \\
    f_{\rho,k}\\
    f_{\omega,k}
\end{bmatrix},
\end{align}
\end{footnotesize}
\noindent where $t_s$ is the discrete time stamp, and $c_\omega$ is the dumping constant value.

Depending on the value of $t_{\alpha,k}$, the status of the cable is taut or slack. Hence, we need to chose the value of $\mathbf{\ddot{p}}_{L,k}$, $f_\rho$ and $f_\omega$, accordingly. Therefore, if $t_{\alpha,k} \leq 0$ (slack mode) then
\begin{align}
    \mathbf{\ddot{p}}_{L,k} = \bm{g}, \quad
    f_{\rho,k} = \boldsymbol{\rho}_k, \quad
    f_{\omega,k} = (1 - t_sc_\omega)\boldsymbol{\omega_{L,k}},
\end{align}
\noindent otherwise, if $t_{\alpha,k} > 0$ (taut mode) then
\begin{footnotesize}
\begin{align}
    \mathbf{\ddot{p}}_{L,k} &= \bm{g} - t_{\alpha,k}\boldsymbol{\rho}_{k},\\
    f_\rho &= \boldsymbol{\rho}_k + t_s (\boldsymbol{\omega}_{L,k} \times \boldsymbol{\rho}_k),\\
    f_\omega &= \boldsymbol{\omega}_{L,k} - \frac{t_s}{L} (\boldsymbol{\rho}_k \times (\mathbf{\ddot{p}}_{Q,k}-\bm{g})) - t_s c_\omega ( \bm{I}_3 - \boldsymbol{\rho}_{k} \boldsymbol{\rho}_{k}^\top) \boldsymbol{\omega_{L,k}},
\end{align}
\end{footnotesize}

Furthermore, the Jacobian $F_k$ can be calculated as
\begin{align}
F_k = 
\begin{bmatrix}
    \bm{I}_3 & t_s \bm{I}_3 & f_{p\rho} & 0\\
    0 & \bm{I}_3 & f_{v\rho} & f_{v\omega} \\
    0 & 0 & \bm{I}_3 + t_s[\boldsymbol{\omega}_{L,k}]_\times & -t_s[\boldsymbol{\rho}_k]_\times\\
    0 & 0 & f_{\omega\rho} &f_{\omega\omega}
\end{bmatrix},
\end{align}
\noindent where $[.]_\times$ is the skew-symmetric matrix, and here again the value of some functions will depend on the value of $t_{\alpha,k}$.

Hence, if $t_{\alpha,k} \leq 0$ (slack mode) then $f_{p\rho} = f_{v\rho} = f_{v\omega} = f_{\omega\rho} = 0$, and $f_{\omega\omega}= ((1 - t_s c_\omega) \bm{I}_3)$, otherwise, if $t_{\alpha,k} > 0$ (taut mode) then
\begin{footnotesize}
\begin{align}
    f_{p\rho}&= \frac{t_s}{2}f_{v\rho},\\
    f_{v\rho} &= t_s\Big(\boldsymbol{\rho}_k (\mathbf{\ddot{p}}_{Q,k}-\bm{g})^\top - (\boldsymbol{\rho}_k^\top (\mathbf{\ddot{p}}_{Q,k}-\bm{g}) - L||\omega||^2)\bm{I}_3\Big),\\
    f_{v\omega} &=  -2Lt_s\Big(\boldsymbol{\rho}_k\boldsymbol{\omega}_{L,k}^\top \Big),\\
    f_{\omega\rho}&= \frac{t_s}{L}[(\boldsymbol{\ddot{P}}_{Q,k}-\bm{g})]_{\times} + t_s c_\omega ((\boldsymbol{\rho}_k^\top \boldsymbol{\omega}_{L,k})\bm{I}_3+\boldsymbol{\rho}_k \boldsymbol{\omega}_{L,k}^\top),\\
   f_{\omega\omega}&= \bm{I}_3 - t_s c_\omega (\bm{I}_3 - \bm{\rho}\bm{\rho}^T).
\end{align}
\end{footnotesize}

Now, we project the state, such that we normalize $\hat{\bm{\rho}}_{k+1|k}$, since it must remain on $\mathbb{S}^2$, and project $\bm{\omega}_{L}$. Thus, if $||\hat{\bm{\rho}}_{k+1|k}||>1e^{-6}$, then
\begin{equation}
    \hat{\bm{\rho}}_{k+1|k} = \frac{\hat{\bm{\rho}}_{k+1|k}}{||\hat{\bm{\rho}}_{k+1|k}||},
\label{reproj1}
\end{equation}
otherwise $\hat{\bm{\rho}}_{k+1|k} = \bm{\rho}_0$.

The projection also influences $\hat{\bm{\omega}}_{L,k+1|k}$, thus
\begin{equation}
    \hat{\bm{\omega}}_{L,k+1|k} = (\bm{I}_3 - \hat{\bm{\rho}}_{k+1|k}\hat{\bm{\rho}}_{k+1|k}^\top)\hat{\bm{\omega}}_{L,k+1|k}.
\label{reproj3}
\end{equation}

We also need to calculate the covariance matrix of the process as follows
\begin{equation}
    \bm{Q}_k = \bm{G}_k \bm{Q}_a \bm{G}_k^\top,
\end{equation}
where $\bm{Q}_k \in \mathbb{R}^{3 \times 3}$ is the process covariance matrix, in which $\bm{Q}_a = \sigma_a^2 \bm{I}_3$, with $\bm{Q}_a \in \mathbb{R}^{3 \times 3}$, and where $\bm{G}_k$ depends on the value of $\hat{t}_{\alpha,k}$, which in turn must be with respect to the estimated states.

Hence, if $\hat{t}_{\alpha,k} \leq 0$ (slack mode) then
\begin{align}
\bm{G}_k = 
    \begin{bmatrix}
     \frac{t_s^2}{2} \hat{\bm{\rho}}_{k+1|k}\bm{I}_3 \\
     \bm{0}_{9\times3}
    \end{bmatrix},
\end{align}
otherwise, if $\hat{t}_{\alpha,k} > 0$ (taut mode) then
\begin{align}
\bm{G}_k = 
    \begin{bmatrix}
     \frac{t_s^2}{2} \hat{\bm{\rho}}_{k+1|k}\hat{\bm{\rho}}_{k+1|k}^\top \\
   t_s\hat{\bm{\rho}}_{k+1|k}\hat{\bm{\rho}}_{k+1|k}^\top \\
    \bm{0}_{3\times3} \\
    -\frac{t_s}{L}[\hat{\bm{\rho}}_{k+1|k}]_\times
    \end{bmatrix}.
\end{align}

The error covariance propagation can then be calculated as
\begin{equation}
    \bm{P}^{+}_k = \bm{F}_k \bm{P}_k \bm{F}_k^\top + \bm{Q}_k.
\end{equation}

We also need to fix the covariance projection with respect to $\hat{\bm{\rho}}_{k+1|k}$. Thus,
\begin{equation}
    \hat{\bm{P}}_{k+1|k} = \bm{B}\hat{\bm{P}}^{+}_k \bm{B}^\top,
\label{reprojP}
\end{equation}
where
\begin{scriptsize}
\begin{align}
    \bm{B} = \begin{bmatrix}
        \bm{I}_3 & 0 & 0 & 0\\
        0 & \bm{I}_3 & 0 & 0\\
        0 & 0 & (\bm{I}_3 - \hat{\bm{\rho}}_{k+1|k}\hat{\bm{\rho}}_{k+1|k}^\top) & 0\\
        0 & 0 & 0 & (\bm{I}_3 - \hat{\bm{\rho}}_{k+1|k}\hat{\bm{\rho}}_{k+1|k}^\top)
        \end{bmatrix}.
\end{align}
\end{scriptsize}

Now, before performing the update, we also need to define the sensor models. With \textbf{no direct cable/load sensors}, our option is to use the quadrotor estimated states from the MEKF. Therefore, we can build up the measurement vector as
\begin{align}
    \bm{z}_k = \begin{bmatrix} \mathbf{p}_{Q,k}^{odom}, \;
    \mathbf{\dot{p}}_{Q,k}^{odom}
    \end{bmatrix}^\top.
\end{align}

The measurement function is then
\begin{align}
    h(\hat{X}_{k+1|k}) =
    \begin{bmatrix}
    \hat{\mathbf{p}}_{L,k+1|k}-L \hat{\bm{\rho}}_{k+1|k} \\
    \hat{\dot{\mathbf{p}}}_{L,k+1|k}-L(\hat{\bm{\omega}}_{L,k+1|k} \times  \hat{\bm{\rho}}_{k+1|k})
    \end{bmatrix}.
\end{align}

Then, we have the innovation such that
\begin{footnotesize}
\begin{align}
    \mathbf{y}_k =
    \begin{bmatrix}
    \mathbf{p}_{Q,k}^{odom} - (\hat{\mathbf{p}}_{L,k+1|k}-L \hat{\bm{\rho}}_{k+1|k}) \\
    \mathbf{\dot{p}}_{Q,k}^{odom} -
    (\hat{\dot{\mathbf{p}}}_{L,k+1|k}-L(\hat{\bm{\omega}}_{L,k+1|k} \times  \hat{\bm{\rho}}_{k+1|k}))
    \end{bmatrix} \in \mathbb{R}^6.
\end{align}
\end{footnotesize}

Finally, the Jacobian H is as follows
\begin{align}
    \bm{H}_k =
    \begin{bmatrix}
    \bm{I}_3 & 0 & -L\bm{I}_3 & 0 \\
    0 & \bm{I}_3 & - L[\bm{\omega}_{L,k}]_\times & L[\bm{\rho}_k]_\times
    \end{bmatrix}.
\end{align}

Now, let's summarize the update as
\begin{align}
    \bm{K}_k &= \hat{\bm{P}}_{k+1|k} \bm{H}_k^\top (\bm{H}_k \hat{\bm{P}}_{k+1|k} \bm{H}_k^\top + \bm{R}_k)^{-1}, \\
    \bm{\xi}_{L,k+1} &= \hat{\bm{\xi}}_{L,k+1|k} + \bm{K}_k \mathbf{y}_k,
\end{align}

We also need, at the end, to perform the re-projection of $\bm{\rho}_{k+1}$ by using equations (\ref{reproj1}-\ref{reproj3}). Furthermore, to calculate the covariance update, we first apply Joseph's form, such that
\begin{small}
\begin{align}
    \bm{P}_{k+1}^{+} &= (\bm{I}_{12} - \bm{K}_k \bm{H}_k) \hat{\bm{P}}_{k+1|k}(\bm{I}_{12} - \bm{K}_k \bm{H}_k)^\top + \bm{K}_k \bm{R}_k \bm{K}_k^\top,\\
    \bm{P}_{k+1} &= 0.5[\bm{P}_{k+1}^{+} (\bm{P}_{k+1}^{+})^\top],
\end{align}
\end{small}
with $\bm{R}_k \in \mathbb{R}^{3 \times 3}$ being the measurement covariance matrix, written as follows
\begin{align}
    \bm{R}_k =
    \begin{bmatrix}
    \sigma_p^2 \bm{I}_3 & 0 \\
    0 & \sigma_v^2 \bm{I}_3
    \end{bmatrix}.
\end{align}

Finally, we also need to apply the same projection into the covariance $\bm{P}_{k+1}$ with respect to $\bm{\rho}_{k+1|k}$ using equation (\ref{reprojP}). 

%% RESULTS %%%%

\begin{figure}[!h]
    \centering
    \includegraphics[width=0.35\textwidth]{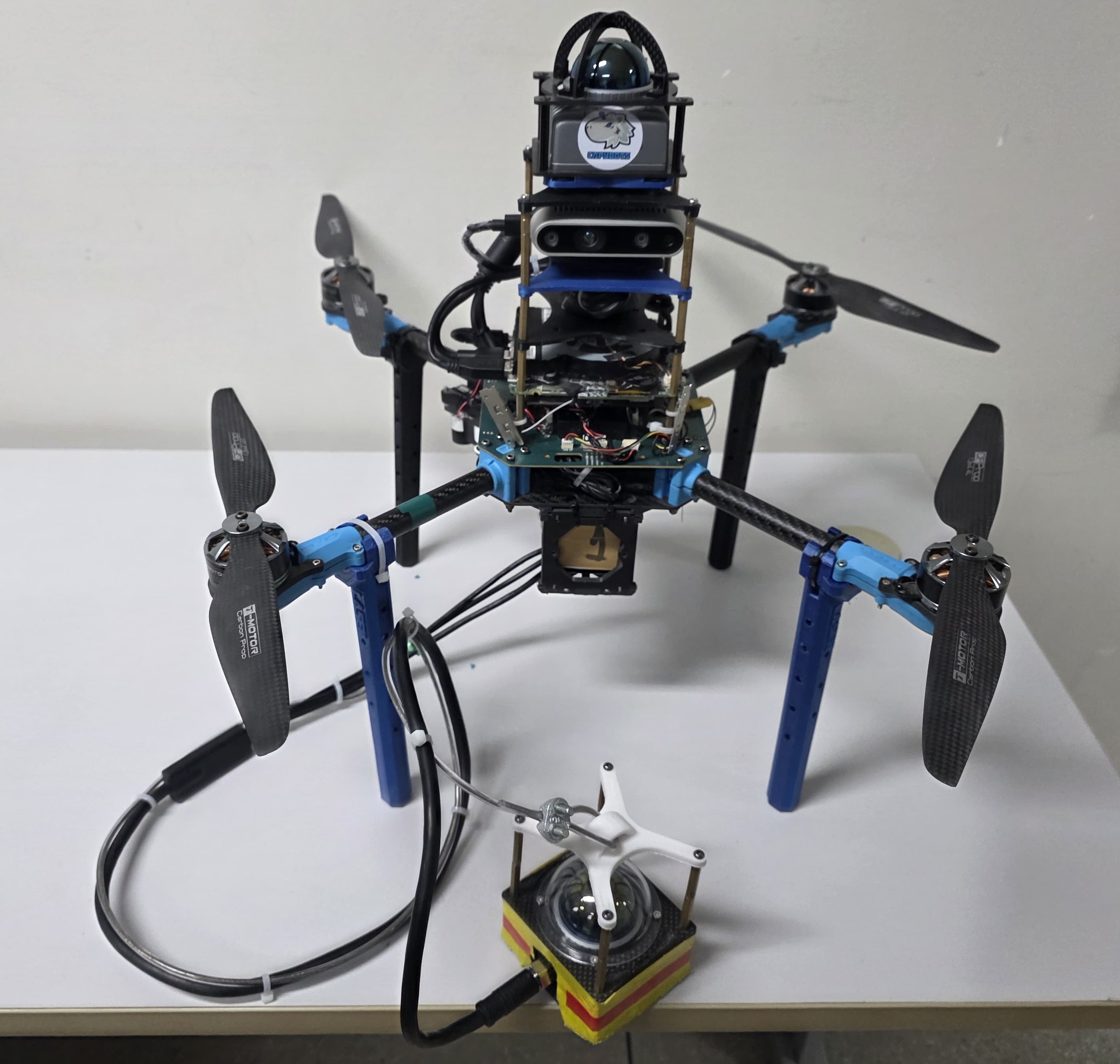}
    \caption{Hardware \ac{uav} platform used in the experiments.}
    \label{fig:realuav}
\end{figure}

\begin{table}[htb]
\centering
\caption{Parameters: Physical Model and Controller.}
\label{tab:parametros_consolidados}
\footnotesize 
\begin{tabular}{lcc}
\toprule
\textbf{Description} & \textbf{Symbol} & \textbf{Value}\\
\midrule
\multicolumn{3}{l}{\textbf{UAV Model Parameters}} \\
\ac{uav} Mass (kg) & $m_Q$  & 2.9 \\
\multirow{3}*{Moments of Inertia (kg$\cdot$m$^2$)}  
        & $I_{xx}$ & 0.03166 \\
        & $I_{yy}$ & 0.03166 \\
        & $I_{zz}$ & 0.04610 \\
Thrust Constant (N$\cdot$s$^2$/rad$^2$)      & $c_{t}$           & 27.6e-6  \\
Torque Constant (N$\cdot$m$\cdot$s$^2$/rad$^2$)      & $c_{q}$           & 16.28e-6 \\
Arm Geometry (m; $^\circ$) & $l_d$, $\beta$        & 0.258, 44.22\\
Quadratic Motor Model    & $a$, $b$  & 0.240572, -0.135153\\
Max Thrust per Motor (N)  & $u_{\text{max}}$         & 17.6 \\
\midrule
\multicolumn{3}{l}{\textbf{Load Model Parameters}} \\
Load Mass (kg) & $m_L$  & 0.5 \\
Cable lenghth (m) & $L_c$  & 1.0 \\
\midrule
\multicolumn{3}{l}{\textbf{Load Estimator}} \\
Dumping Constant & $c_\omega$  & 0.4 \\
Acceleration Variance & $\sigma^2_a$  & 0.15 \\
Variances for Measurement Matrix & $\sigma^2_p, \sigma^2_v$  & 0.0025; 0.0025 \\
\midrule
\multicolumn{3}{l}{\textbf{\ac{nmpc} Controller}} \\
Prediction Horizon (-; s)    & $N, dt$  & 20; 0.05 \\
UAV State Error Weight  & $\mathbf{S_Q}$ & diag($100; 80; 0.3; 1.0$)\\
Load State Error Weight  & $\mathbf{S_L}$ & diag($0.3; 0.3; 20.0; 0.3$)\\
Control Effort Weight          & $\mathbf{C}$      & $I_8$\\
\bottomrule
\end{tabular}
\end{table}

\begin{figure*}[h!]
    \centering
    \begin{subfigure}[b]{0.45\textwidth}
        \centering
        \includegraphics[width=\linewidth]{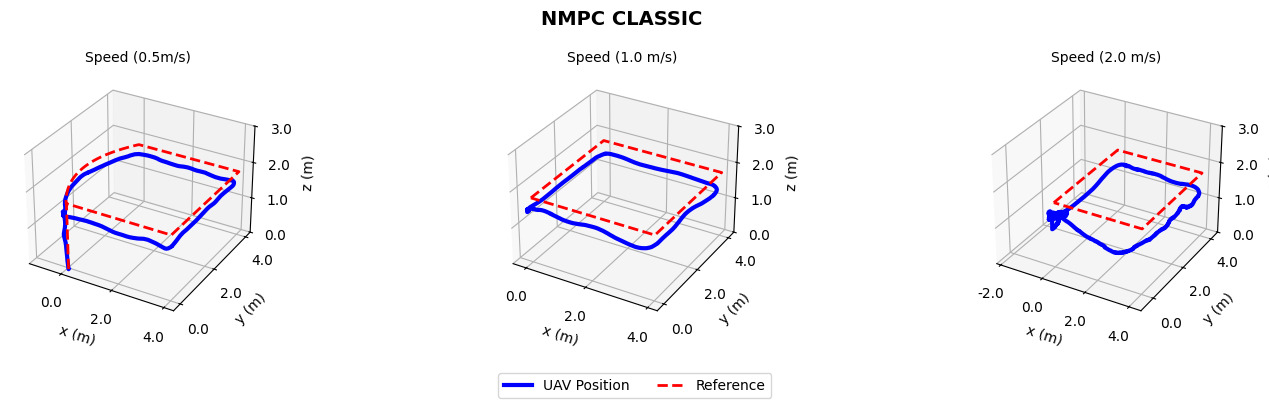}
        % \caption{NMPC}
    \end{subfigure}
    \hfill
    \begin{subfigure}[b]{0.45\textwidth}
        \centering
        \includegraphics[width=\linewidth]{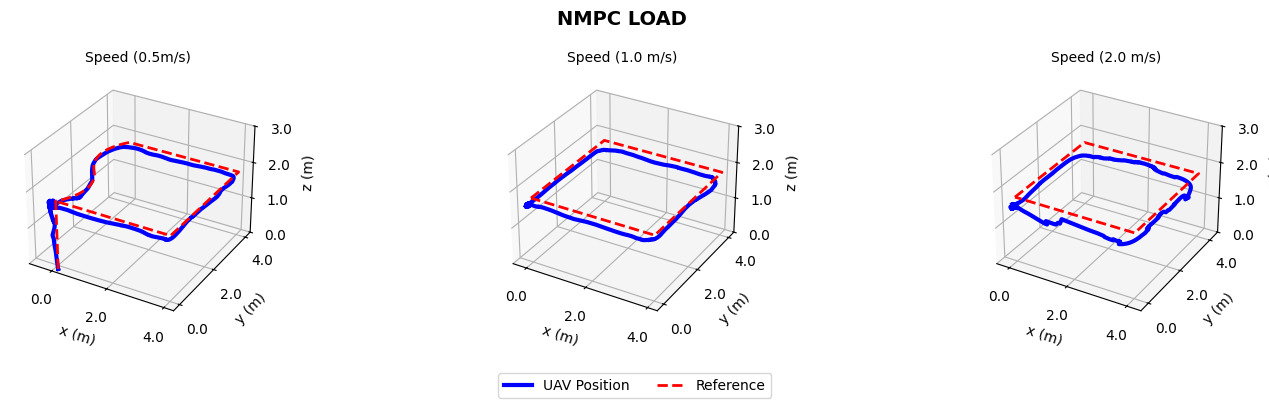}
        % \caption{NMPC-Load}
    \end{subfigure}
    \begin{subfigure}[b]{0.45\textwidth} 
        \centering 
        \includegraphics[width=\linewidth]{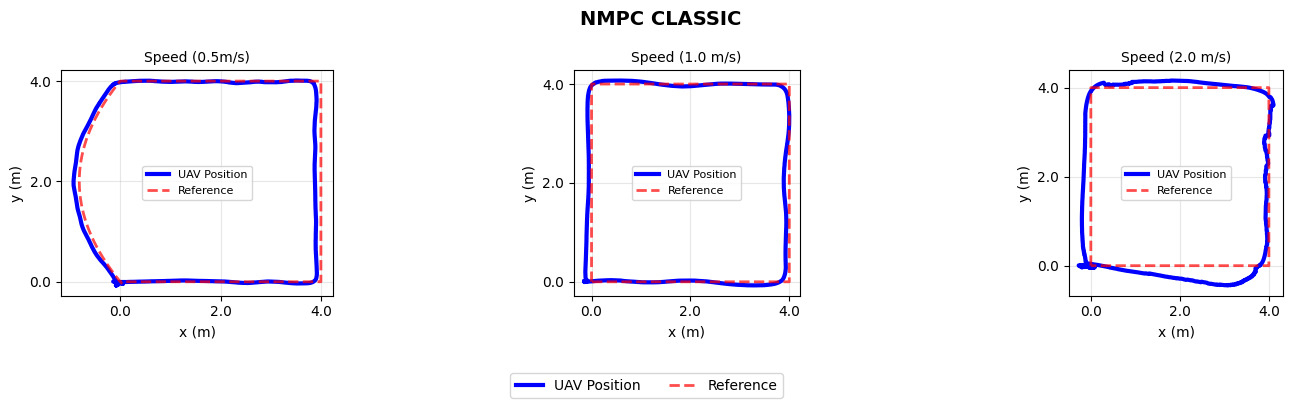} 
        % \caption{NMPC} 
    \end{subfigure} 
    \hfill 
    \begin{subfigure}[b]{0.45\textwidth} 
        \centering 
        \includegraphics[width=\linewidth]{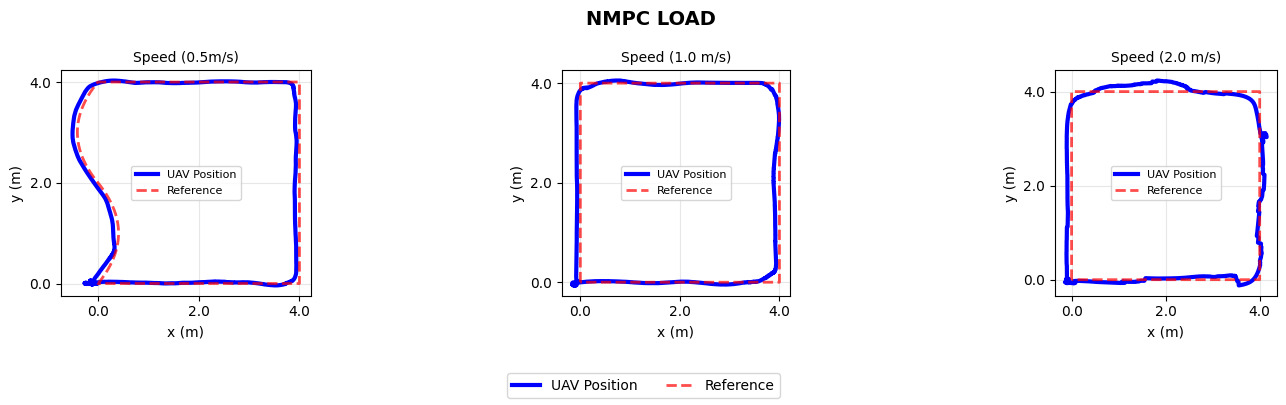} 
        % \caption{NMPC-Load} 
    \end{subfigure} 
    \caption{Executed trajectories with the classic \ac{nmpc} (left) and our proposed NMPC-Load (right) with trajectory speeds of 0.5 m/s, 1.0 m/s, and 2.0 m/s, respectively}
    \label{fig:traj_top_both}
\end{figure*}

\section{Results}

To analyze the robustness of the controllers, we conducted real-robot experiments with the explicit objective of stressing the UAV-payload system dynamics (see Fig. \ref{fig:realuav} and Video\footnote{Video of the Experiment: %\url{https://youtu.be/7_4bGrbSdaI}
}). The parameters used in the experiment are in Table \ref{tab:parametros_consolidados}. Regarding computational capacity and control-loop efficiency, the experiment was conducted on an \ac{uav} equipped with an onboard computer featuring a 12th-generation Intel Core i7 processor and $32~\text{GB}$ of DDR4 RAM. It was verified that the processing load during flight was approximately $8\%$ for this project's modules, thereby validating the computational efficiency of the ACADOS framework used to solve the \ac{nmpc} \ac{ocp}. The \ac{uav} was equipped with a RealSense camera, which was used in the Visual Inertial Odometry algorithm to estimate the UAV's state during flight.

In this work, we compare our approach (\ac{nmpc} with a complete load model, or NMPC-Load) with the classic \ac{nmpc} approach by conducting the following experiment: an \ac{uav} transported a payload along a square trajectory via a massless cable at a constant altitude of 1.8 m. On the payload, a Livox Mid 360 sensor was installed, \textbf{only to measure the payload's ground truth in the XY-axis}. The waypoints that define the route were $P_1=[0.0, 0.0, 1.8]$, $P_2=[4.0, 0.0, 1.8]$, $P_3=[4.0, 4.0, 1.8]$, $P_4=[0.0, 4.0, 1.8]$, and $P_2=[0.0, 0.0, 1.8]$. The executed trajectory contains four straight segments and four abrupt 90° changes of direction, introducing significant transverse accelerations whenever the \ac{uav} executes each turn. Under suspended-load conditions, these lateral accelerations are particularly relevant, as they induce large cable deflections and excite natural oscillatory modes of the load. Thus, this simulation scenario was selected to evaluate each controller's ability to handle load-induced oscillations, reject dynamic disturbances during rapid changes of direction, track the reference trajectory under heavy load, and assess the system's sensitivity to UAV-payload dynamic coupling. Finally, the \ac{uav} performed this trajectory three times, each time with increasing speed of 0.5 m/s, 1.0 m/s, and 2.0 m/s. The increase in trajectory reference speed significantly intensifies dynamic coupling, increasing the amplitude and persistence of payload oscillations and posing additional challenges for the controller. 

Figure \ref{fig:traj_top_both} compares the classical \ac{nmpc} approach with our NMPC-Load approach by showing the executed trajectories (top view) for both approaches. We also evaluate both controllers using Root Mean Square Error (RMSE) and its Standard Deviation (STD), even when the estimated and reference trajectories are sampled at different times \cite{Zhang18iros}. The method is based on evaluating sub-trajectories, enabling distinction between local errors, dominated by rapid oscillations and instantaneous variations, and accumulated errors, which are more sensitive to the controller's ability to maintain stability over extended paths. By analyzing both Figure \ref{fig:traj_top_both} and the results presented in Table \ref{tab:resultados_erro_uav_real}, we can perceive that our proposed \textbf{NMPC-Load} controller outperforms the \textbf{NMPC} controller in all metrics.
\begin{table}[h!]
\centering
\caption{Relative errors for the square trajectory.}
\label{tab:resultados_erro_uav_real}
\begin{tabular}{lcc}
\toprule
\textbf{Metrics} & \textbf{NMPC} & \textbf{NMPC-Load} \\
\midrule
RMSE Drone  & 0.5251  & 0.3732 \\
STD Drone    & 0.3384  & 0.1928 \\
RMSE Payload    & 0.8034  & 0.7797 \\
STD Payload    & 0.5444  & 0.4719 \\
\bottomrule
\end{tabular}
\end{table}

\begin{figure}[!t]
    \centering
    \includegraphics[width=0.45\textwidth]{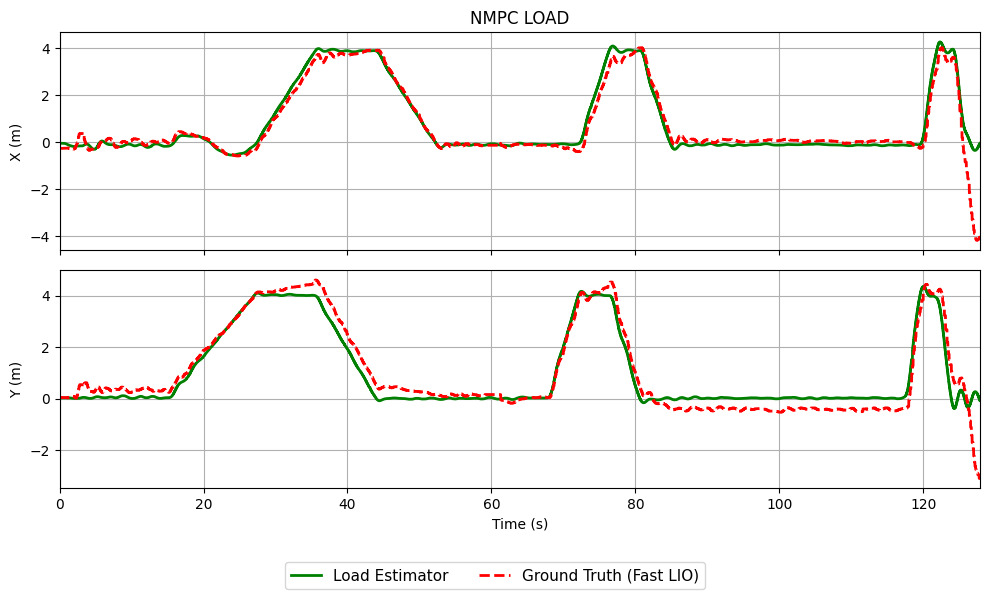}
    \caption{Schematic of a \ac{uav} carrying a payload via a cable with axis convention.}
    \label{fig:loadestimation}
\end{figure}

The second reason for this performance is due to our proposed load estimator. Fig. \ref{fig:loadestimation} shows the behavior of our estimation compared with the ground-truth data of the payload. It is worth highlighting that the estimator was able to, \textbf{without any external sensor on the cable or the load} (only using the UAV's IMU data), correctly estimate the load's position during the flight.

These results show that, even in short sections, the controller that incorporates load dynamics is more effective in suppressing oscillations induced by pendulum motion. Furthermore, the NMPC-Load control incorporates the complete load system model, reducing accumulated errors and ensuring accuracy during the tracking of challenging trajectories.

\section{Conclusion}

In this work, we hope to have pushed the boundaries of the state-of-the-art of sensorless estimation and control for agile
cable-suspended aerial payload transportation. By introducing a method to model both the external force induced by the load and the \ac{uav} within the Newton-Euler formulation through the principle of Udwadia-Kalaba, we were able the use of both models within a \ac{nmpc} control system in a more straightforward manner. Furthermore, the load model was enhanced with a load estimator that only used the UAV's IMU as control input and the \ac{uav} state estimation as measurement, resulting in a numerically robust and physically coherent controller. No additional sensor was needed. The results showed that taking these considerations into account enabled better load-carrying performance (i.e., reduced load swing, improved motion tracking, etc.) while also enabling agile flight.

\bibliographystyle{IEEEtran}
\bibliography{bibliography}
\newpage

\vfill

\end{document}